\documentclass{sig-alternate-2013}
\usepackage{graphicx}
\usepackage{subfigure}
\usepackage{amsmath}
\usepackage{tabularx}
\usepackage{amssymb}
\usepackage{mathrsfs}
\usepackage{algorithm}
\usepackage{algorithmic}
\usepackage{multirow}
\usepackage{bm}

\newfont{\mycrnotice}{ptmr8t at 7pt}
\newfont{\myconfname}{ptmri8t at 7pt}
\let\crnotice\mycrnotice%
\let\confname\myconfname%

\permission{Permission to make digital or hard copies of all or part of this work for personal or classroom use is granted without fee provided that copies are not made or distributed for profit or commercial advantage and that copies bear this notice and the full citation on the first page. Copyrights for components of this work owned by others than ACM must be honored. Abstracting with credit is permitted. To copy otherwise, or republish, to post on servers or to redistribute to lists, requires prior specific permission and/or a fee. Request permissions from Permissions@acm.org.}
\conferenceinfo{MM'15,}{October 26--30, 2015, Brisbane, Australia.} 
\copyrightetc{\copyright~2015 ACM. ISBN \the\acmcopyr}
\crdata{978-1-4503-3459-4/15/10\ ...\$15.00.\\
DOI: http://dx.doi.org/10.1145/XXX.XXXXXXX}

\begin{document}

%

\title{DeepFont: Identify Your Font from An Image}

\author{
\alignauthor Zhangyang Wang$^1$\ \ \ \ \ Jianchao Yang$^3$\ \ \ \ \ Hailin Jin$^2$\ \ \ \ \ Eli Shechtman$^2$\ \ \ \ \ Aseem Agarwala $^4$ Jonathan Brandt $^2$\ \ \ \ \  Thomas S. Huang $^1$ \\
\affaddr{$^1$University of Illinois at Urbana-Champaign}\\
\affaddr{$^2$Adobe Research}\\
\affaddr{$^3$Snapchat Inc \:}
\affaddr{$^4$Google Inc}\\
\email{\tt \{zwang119, t-huang1\}@illinois.edu, jianchao.yang@snapchat.com, \{hljin, elishe, jbrandt\}@adobe.com, aseem@agarwala.org}
}

\maketitle
\begin{abstract}
As font is one of the core design concepts, automatic font identification and similar font suggestion from an image or photo has been on the wish list of many designers. We study the Visual Font Recognition (VFR) problem \cite{LFE}, and advance the state-of-the-art remarkably by developing the \textit{DeepFont} system. First of all, we build up the first available large-scale VFR dataset, named \textit{AdobeVFR}, consisting of both labeled synthetic data and partially labeled real-world data. Next, to combat the domain mismatch between available training and testing data, we introduce a Convolutional Neural Network (CNN) decomposition approach, using a domain adaptation technique based on a Stacked Convolutional Auto-Encoder (SCAE) that exploits a large corpus of unlabeled real-world text images combined with synthetic data preprocessed in a specific way. Moreover, we study a novel learning-based model compression approach, in order to reduce the DeepFont model size without sacrificing its performance. The DeepFont system achieves an accuracy of higher than 80\% (top-5) on our collected dataset, and also produces a good font similarity measure for font selection and suggestion. We also achieve around 6 times compression of the model without any visible loss of recognition accuracy. 
\end{abstract}




\section{Introduction}
Typography is fundamental to graphic design. Graphic designers have the desire to identify the fonts they encounter in daily life for later use. While they might take a photo of the text of a particularly interesting font and seek out an expert to identify the font, the manual identification process is extremely tedious and error-prone. Several websites allow users to search and recognize fonts by font similarity, including Identifont, MyFonts, WhatTheFont, and Fontspring. All of them rely on tedious humans interactions and high-quality manual pre-processing of images, and the accuracies are still unsatisfactory. On the other hand, the majority of font selection interfaces in existing softwares are simple linear lists, while exhaustively exploring the entire space of fonts using an alphabetical listing is unrealistic for most users. 

Effective automatic font identification from an image or photo could greatly ease the above difficulties, and facilitate font organization and selection during the design process. Such a Visual Font Recognition (VFR) problem is inherently difficult, as pointed out in \cite{LFE}, due to the huge space of possible fonts (online repositories provide hundreds of thousands), the dynamic and open-ended properties of font classes, and the very subtle and character-dependent difference among fonts (letter endings, weights, slopes, etc.). More importantly, while the popular machine learning techniques are data-driven, collecting real-world data for a large collection of font classes turns out to be extremely difficult. Most attainable real-world text images do not have font label information, while the error-prone font labeling task requires font expertise that is out of reach of most people. The few previous approaches \cite{3,4,5,6,7,8} are mostly from the document analysis standpoint, which only focus on a small number of font classes, and are highly sensitive to noise, blur, perspective distortions, and complex backgrounds. In \cite{LFE} the authors proposed a large-scale, learning-based solution without dependence on character segmentation or OCR. The core algorithm is built on local feature embedding, local feature metric learning and max-margin template selection. However, their results suggest that the robustness to real-world variations is unsatisfactory, and a higher recognition accuracy is still demanded.

\begin{figure*}[htbp]
\centering
\begin{minipage}{0.380\textwidth}
\centering \subfigure[] {
\includegraphics[width=\textwidth]{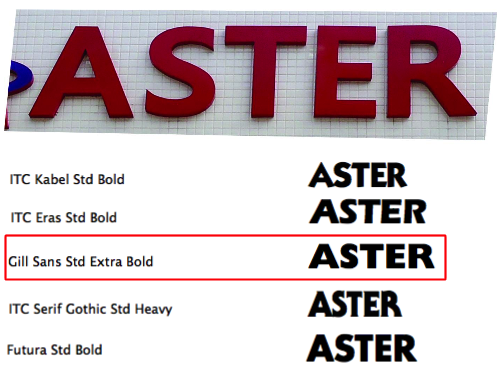}
}\end{minipage}
\begin{minipage}{0.505\textwidth}
\centering \subfigure[] {
\includegraphics[width=\textwidth]{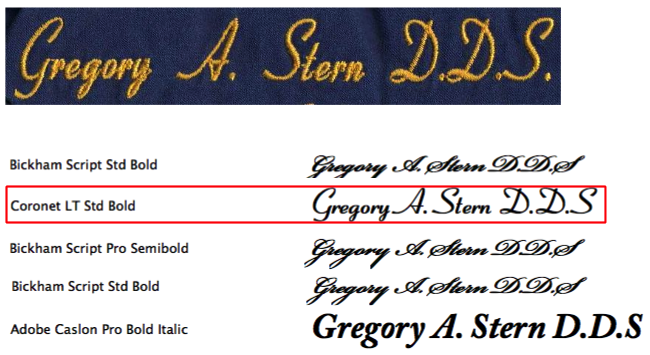}
}\end{minipage}
\\
\begin{minipage}{0.82\textwidth}
\centering \subfigure [] {
\includegraphics[width=\textwidth]{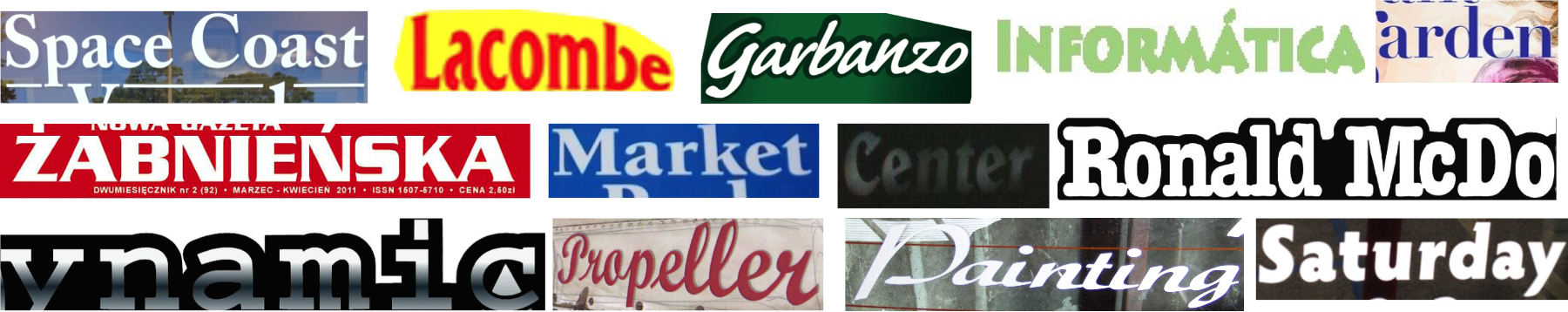}
}\end{minipage}
\caption{(a) (b) Successful VFR examples with the DeepFont system. The top row are query images from VFR\_real\_test  dataset. Below each query, the results (left column: font classes; right column: images rendered with the corresponding font classes) are listed in a high-to-low order in term of likelihoods. The correct results are marked by the red boxes. (c) More correctly recognized real-world images with DeepFont.}
\label{fig:show}
\end{figure*}

Inspired by the great success achieved by deep learning models \cite{imagenet} in many other computer vision tasks, we develop a VFR system for the Roman alphabets, based on the Convolutional neural networks (CNN), named \textit{DeepFont}. Without any dependence on character segmentation or content text, the DeepFont system obtains an impressive performance on our collected large real-word dataset, covering an extensive variety of font categories. Our technical contributions are listed below:
\begin{itemize}
\item \textbf{AdobeVFR Dataset} A large set of \textit{labeled real-world} images as well as a large corpus of unlabeled real-world data are collected for both training and testing, which is the first of its kind and is publicly released soon. We also leverage a large training corpus of labeled synthetic data augmented in a specific way.
\item \textbf{Domain Adapted CNN} It is very easy to generate lots of rendered font examples but very hard to obtain labeled real-world images for supervised training. This real-to-synthetic domain gap caused poor generalization to new real data in previous VFR methods \cite{LFE}. We address this \textit{domain mismatch} problem by leveraging synthetic data to obtain effective classification features, while introducing a domain adaptation technique based on Stacked Convolutional Auto-Encoder (SCAE) with the help of unlabeled real-world data.
\item \textbf{Learning-based Model Compression} We introduce a novel learning-based approach to obtain a \textit{losslessly compressible model}, for a high compression ratio without sacrificing its performance. An exact low-rank constraint is enforced on the targeted weight matrix.
\end{itemize}
Fig. \ref{fig:show} shows successful VFR examples using DeepFont. In (a)(b), given the real-world query images, top-5 font recognition results are listed, within which the ground truth font classes are marked out\footnote{Note that the texts are input manually for rendering purposes only. The font recognition process does not need any content information.}. More real-world examples are displayed in (c). Although accompanied with high levels of background clutters, size and ratio variations, as well as perspective distortions, they are all correctly recognized by the DeepFont system.  

 \begin{table}[t]
 \begin{center}
 \caption{Comparison of All VFR Datasets}
 \label{data}
 \vspace{0.1em}
 \begin{tabular}{|c|c|c|c|c|c|}
 \hline
 Dataset name& Source & Label? & Purpose & Size & Class  \\
 \hline
 VFRWild325 \cite{LFE} & Real &  Y & Test & 325 & 93 \\
 \hline
 VFR\_real\_test & Real &  Y& Test &  4, 384 & 617 \\
 \hline
 VFR\_real\_u & Real & N & Train &  197, 396 & /\ \\
 \hline
 VFR\_syn\_train & Syn & Y &  Train & 2,383, 000& 2, 383\\
 \hline
 VFR\_syn\_val & Syn & Y & Test & 238, 300 & 2, 383 \\
 \hline
 \end{tabular}
 \end{center}
 \end{table}
 
\section{Dataset}

\subsection{Domain Mismatch between Synthetic and Real-World Data}

To apply machine learning to VFR problem, we require realistic text images with ground truth font labels.  However, such data is scarce and expensive to obtain. Moreover, the training data requirement is vast, since there are hundreds of thousands of fonts in use for Roman characters alone. One way to overcome the training data challenge is to synthesize the training set by rendering text fragments for all the necessary fonts. However, to attain effective recognition models with this strategy, we must face the domain mismatch between synthetic and real-world text images \cite{LFE}.  For example, it is common for designers to edit the spacing, aspect ratio or alignment of text arbitrarily, to make the text fit other design components. The result is that characters in real-world images are spaced, stretched and distorted in numerous ways.  For example, Fig.\ \ref{fig:per} (a) and (b) depict typical examples of \textit{character spacing} and \textit{aspect ratio} differences between (standard rendered) synthetic and real-world images.  Other perturbations, such as background clutter, perspective distortion, noise, and blur, are also ubiquitous.

\subsection{The AdobeVFR Dataset}

Collecting and labeling real-world examples is notoriously hard and thus a labeled real-world dataset has been absent for long. A small dataset \textit{VFRWild325} was collected in \cite{LFE}, consisting of 325 real-world text images and 93 classes. However, the small size puts its effectiveness in jeopardy.

Chen et. al. in \cite{LFE} selected 2,420 font classes to work on. We remove some script classes, ending up with a total of 2,383 font classes.  We collected 201,780 text images from various typography forums, where people post these images seeking help from experts to identify the fonts. Most of them come with hand-annotated font labels which may be inaccurate. Unfortunately, only a very small portion of them fall into our list of 2,383 fonts. All images are first converted into gray scale. Those images with our target class labels are then selected and inspected by independent experts if their labels are correct. Images with verified labels are then manually cropped with tight bounding boxes and normalized proportionally in size, to be with the identical height of 105 pixels. Finally, we obtain 4,384 real-world test images with reliable labels, covering 617 classes (out of 2,383). Compared to the synthetic data, these images typically have much larger appearance variations caused by scaling, background clutter, lighting, noise, perspective distortions, and compression artifacts. Removing the 4,384 labeled images from the full set, we are left with 197,396 unlabeled real-world images which we denote as VFR\_real\_u.

To create a sufficiently large set of synthetic training data, we follow the same way in \cite{LFE} to render long English words sampled from a large corpus, and generate tightly cropped, gray-scale, and size-normalized text images. For each class, we assign 1,000 images for training, and 100 for validation, which are denoted as VFR\_syn\_train and  VFR\_syn\_val, respectively. The entire \textbf{AdobeVFR dataset}, consisting of  VFR\_real \_test,  VFR\_real\_u, VFR\_syn\_train and  VFR\_syn\_val, are made publicly available\footnote{http://www.atlaswang.com/deepfont.html}.

The AdobeVFR dataset is the first large-scale benchmark set consisting of both synthetic and real-world text images, for the task of font recognition. To our best knowledge, so far VFR\_real\_test is the largest available set of real-world text images with reliable font label information (12.5 times larger than VFRWild325). The AdobeVFR dataset is super fine-grain, with highly subtle categorical variations, leading itself to a new challenging dataset for object recognition. Moreover, the substantial mismatch between synthetic and real-world data makes the AdobeVFR dataset an ideal subject for general domain adaption and transfer learning research. It also promotes the new problem area of understanding design styles with deep learning.

\subsection{Synthetic Data Augmentation: A First Step to Reduce the Mismatch}

\begin{figure}[htbp]
\centering
\begin{minipage}{0.20\textwidth}
\centering \subfigure[] {
\includegraphics[width=\textwidth]{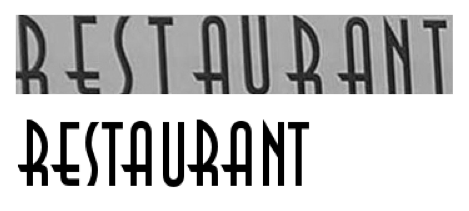}
}\end{minipage}
\begin{minipage}{0.20\textwidth}
\centering \subfigure [] {
\includegraphics[width=\textwidth]{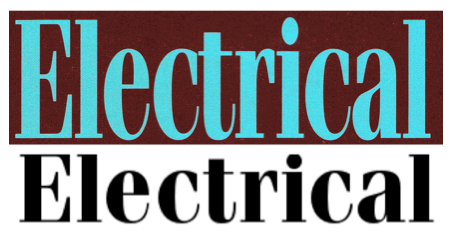}
}\end{minipage}
\caption{(a) the different characters spacings between a pair of synthetic and real-world images. (b) the different aspect ratio between a pair of synthetic and real-world image}
\label{fig:per}
\end{figure}

Before feeding synthetic data into model training, it is popular to artificially augment training data using label-preserving transformations to reduce overfitting. In \cite{imagenet}, the authors applied image translations and horizontal reflections to the training images, as well as altering the intensities of their RGB channels. The authors in \cite{LFE} added moderate distortions and corruptions to the synthetic text images:
\begin{itemize}
\item \textbf{1. Noise:} a small Gaussian noise with zero mean and standard deviation 3 is added to input
\item \textbf{2. Blur:} a random Gaussian blur with standard deviation from 2.5 to 3.5 is added to input
\item \textbf{3. Perspective Rotation:} a randomly-parameterized affine transformation is added to input
\item \textbf{4. Shading:} the input background is filled with a gradient in illumination.
\end{itemize}
The above augmentations cover standard perturbations for general images, and are adopted by us. However, as a very particular type of images, text images have various real-world appearances caused by specific handlings. Based on the observations in Fig. \ref{fig:per} , we identify two additional font-specific augmentation steps to our training data:
\begin{itemize}
\item \textbf{5. Variable Character Spacing: }  when rendering each synthetic image, we set the character spacing (by pixel) to be a Gaussian random variable of mean 10 and standard deviation 40, bounded by [0, 50].
\item \textbf{6. Variable Aspect Ratio: } Before cropping each image into a input patch, the image, with heigh fixed, is squeezed in width by a random ratio, drawn from a uniform distribution between $\frac{5}{6}$ and $\frac{7}{6}$. 
\end{itemize}
Note that these steps are not useful for the method in \cite{LFE} because it exploits very localized features. However, as we show in our experiments, these steps lead to significant performance improvements in our DeepFont system.  Overall, our data augmentation includes steps 1-6.

\begin{figure}[htbp]
\centering
\begin{minipage}{0.20\textwidth}
\centering \subfigure[Synthetic, none] {
\includegraphics[width=\textwidth]{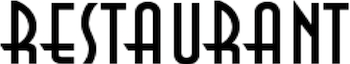}
}
\centering \subfigure[Synthetic, 1-4] {
\includegraphics[width=\textwidth]{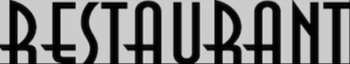}
}\end{minipage}
\begin{minipage}{0.20\textwidth}
\centering \subfigure[Synthetic, 5-6] {
\includegraphics[width=\textwidth]{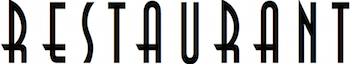}
}
\centering \subfigure[Synthetic, 1-6] {
\includegraphics[width=\textwidth]{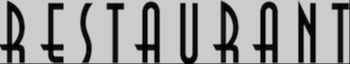}
}\end{minipage}
\begin{minipage}{0.35\textwidth}
\centering \subfigure[Relative CNN layer-wise responses] {
\includegraphics[width=\textwidth]{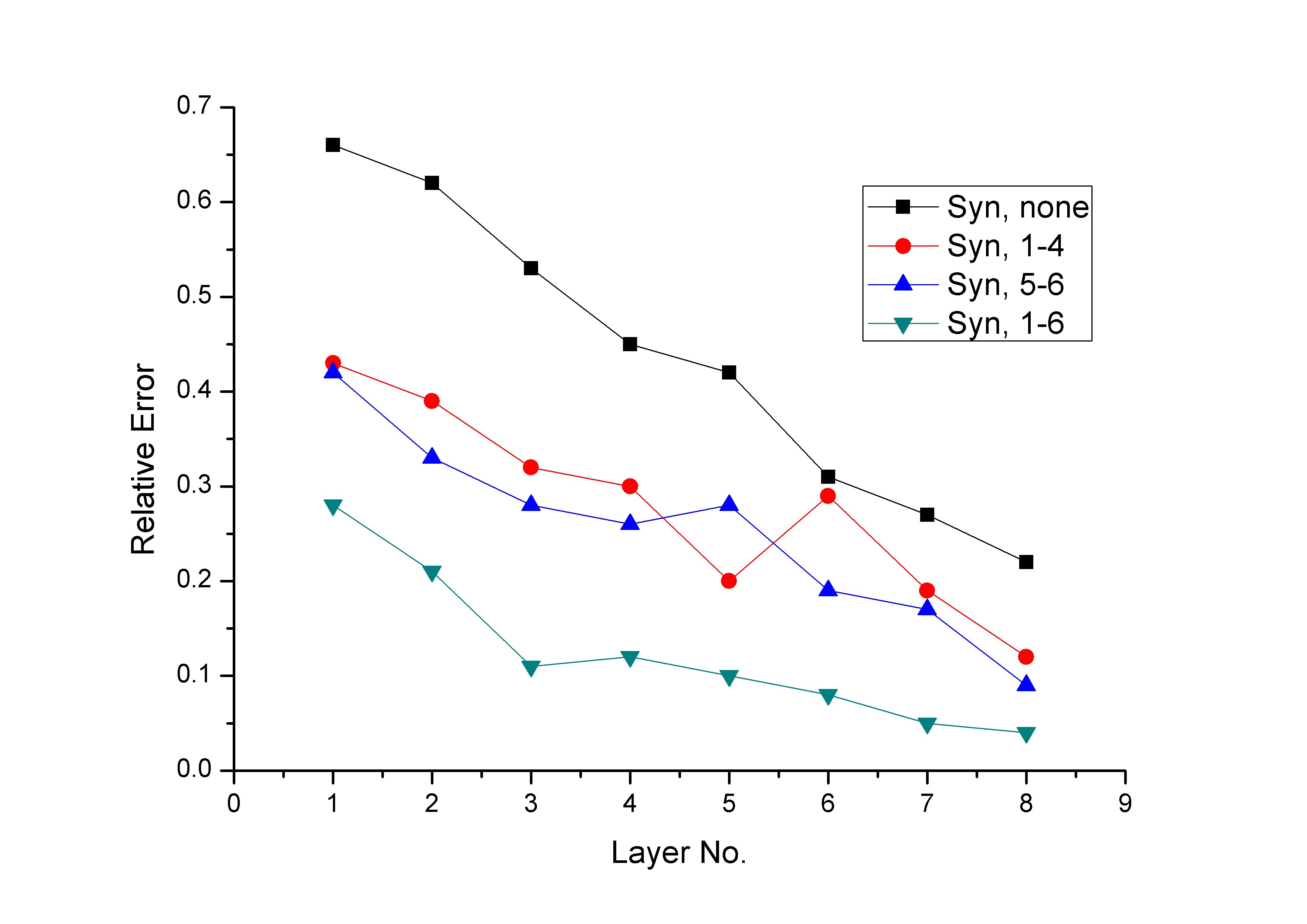}
}\end{minipage}
\caption{The effects of data augmentation steps. (a)-(d): synthetic images of the same text but with different data augmentation ways. (e) compares relative differences of (a)-(d) with the real-world image Fig. \ref{fig:per} (a), in the measure of layer-wise network activations through the same DeepFont model.}
\label{fig:pre}
\end{figure}

\begin{figure}[htbp]
\centering
\begin{minipage}{0.49\textwidth}
\centering {
\includegraphics[width=\textwidth]{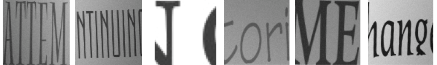}
}\end{minipage}
\caption{Examples of synthetic training $105 \times105$ patches after pre-processing steps 1-6.}
\label{fig:patch}
\end{figure}

To leave a visual impression, we take the real-world image Fig. \ref{fig:per} (a), and synthesize a series of images in Fig. \ref{fig:pre}, all with the same text but with different data augmentation ways. Specially, (a) is synthesized with no data augmentation; (b) is (a) with standard augmentation 1-4 added; (c) is synthesized with spacing and aspect ratio customized to be identical to those of Fig. \ref{fig:per} (a); (d) adds standard augmentation 1-4 to (c). We input images (a)-(d) through the trained DeepFont model. For each image, we compare its layer-wise activations with those of the real image Fig. \ref{fig:per} (a) feeding through the same model, by calculating the normalized MSEs. Fig. \ref{fig:pre} (e) shows that those augmentations, especially the spacing and aspect ratio changes, reduce the gap between the feature hierarchies of real-world and synthetic data to a large extent.
A few synthetic patches after full data augmentation 1-6 are displayed in Fig. \ref{fig:patch}. It is observable that they possess a much more visually similar appearance to real-world data.

\section{Domain Adapted CNN}

\subsection{Domain Adaptation by CNN Decomposition and SCAE}
Despite that data augmentations are helpful to reduce the domain mismatch, enumerating all possible real-world degradations is impossible, and may further introduce degradation bias in training. In the section, we propose a learning framework to leverage both synthetic and real-world data, using multi-layer CNN decomposition and SCAE-based domain adaptation. Our approach extends the domain adaptation method in \cite{sentiment} to extract low-level features that represent both the synthetic and real-world data. We employs a Convolutional Neural Network (CNN) architecture, which is further decomposed into two sub-networks: a "shared" low-level sub-network which is learned from the composite set of synthetic and real-world data, and a high-level sub-network that learns a deep classifier from the low-level features. 

\begin{figure*}[htbp]
\centering
\begin{minipage}{0.95\textwidth}
\centering {
\includegraphics[width=\textwidth]{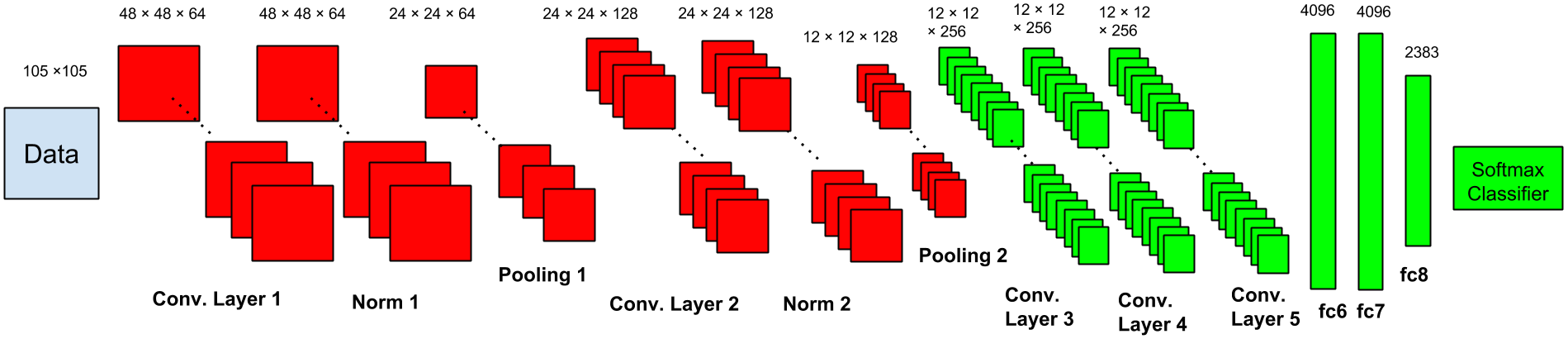}
}\end{minipage}
\caption{The CNN architecture in the DeepFont system, and its decomposition marked by different colors ($N$=8, $K$=2).}
\label{fig:CNN}
\end{figure*}

The basic CNN architecture is similar to the popular ImageNet structure \cite{imagenet}, as in Fig. \ref{fig:CNN}. The numbers along with the network pipeline specify the dimensions of outputs of corresponding layers.  The input is a $105 \times 105$ patch sampled from a "normalized" image. Since a square window may not capture sufficient discriminative local structures, and is unlikely to catch high-level combinational features when two or more graphemes or letters are joined as a single glyph (e.g., ligatures), we introduce a \textbf{squeezing} operation \footnote{Note squeezing is independent from the variable aspect ratio operation introduced in Section 2.3, as they are for different purposes.}, that scales the width of the height-normalized image to be of a constant ratio relative to the height (2.5 in all our experiments). Note that the squeezing operation is equivalent to producing ``long'' rectangular input patches. 

When the CNN model is trained fully on a synthetic dataset, it witnesses a significant performance drop when testing on real-world data, compared to when applied to another synthetic validation set. This also happens with other models such as in \cite{LFE}, which uses training and testing sets of similar properties to ours. It alludes to discrepancies between the distributions of synthetic and real-world examples. we propose to decompose the $N$ CNN layers into two sub-networks to be learned sequentially:
\begin{itemize}
\item \textbf{Unsupervised cross-domain sub-network $\mathbf{C_u}$}, which consists of the first $K$ layers of CNN. It accounts for extracting low-level visual features shared by both synthetic and real-world data domains. $\mathbf{C_u}$ will be trained in a unsupervised way, using unlabeled data from both domains. It constitutes the crucial step that further minimizes the low-level feature gap, beyond the previous data augmentation efforts.
\item \textbf{Supervised domain-specific sub-network $\mathbf{C_s}$}, which consists of the remaining $N-K$ layers. It accounts for learning higher-level discriminative features for classification, based on the shared features from $\mathbf{C_u}$. $\mathbf{C_s}$ will be trained in a supervised way, using labeled data from the synthetic domain only.
\end{itemize}
We show an example of the proposed CNN decomposition in Fig. \ref{fig:CNN}. The $\mathbf{C_u}$ and $\mathbf{C_s}$ parts are marked by red and green colors, respectively, with $N=8$ and $K=2$. Note that the low-level shared features are implied to be independent of class labels. Therefore in order to address the open-ended problem of font classes, one may keep re-using the $\mathbf{C_u}$ sub-network, and only re-train the $\mathbf{C_s}$ part.

\begin{figure}[htbp]
\centering
\begin{minipage}{0.49\textwidth}
\centering {
\includegraphics[width=\textwidth]{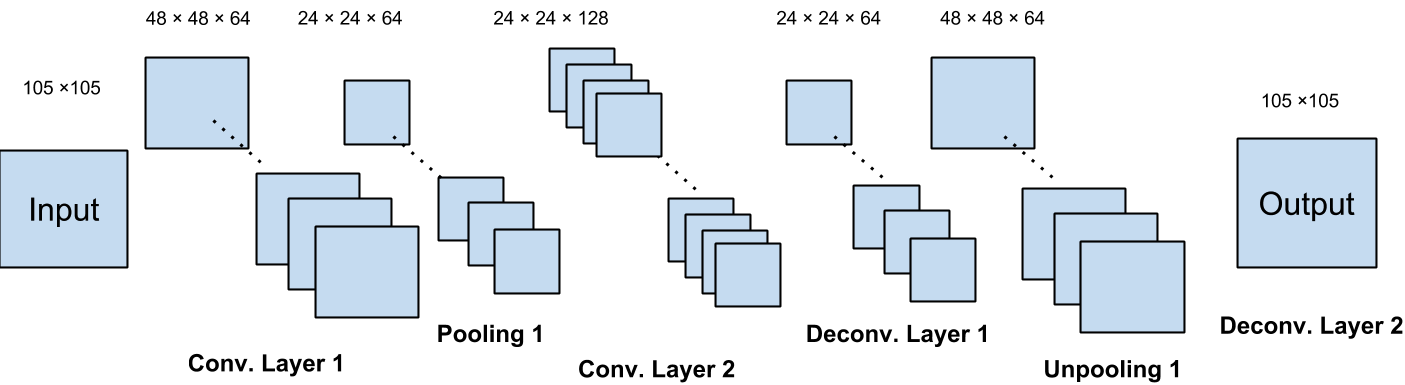}
}\end{minipage}
\caption{The Stacked Convolutional Auto-Encoder (SCAE) architecture.}
\label{fig:CAE}
\end{figure}

\noindent  \textbf{Learning $\mathbf{C_u}$ from SCAE} Representative unsupervised feature learning methods, such as the Auto-Encoder and the Denoising Auto-Encoder, perform a greedy layer-wise pre-training of weights using unlabeled data alone followed by supervised fine-tuning (\cite{semi3}). However, they rely mostly on fully-connected models and ignore the 2D image structure.  In \cite{SCAE}, a Convolutional Auto-Encoder (CAE) was proposed to learn non-trivial features using a hierarchical unsupervised feature extractor that scales well to high-dimensional inputs. The CAE architecture is intuitively similar to the the conventional auto-encoders in \cite{AE}, except for that their weights are shared among all locations in the input, preserving spatial locality. CAEs can be stacked to form a deep hierarchy called the Stacked Convolutional Auto-Encoder (SCAE), where each layer receives its input from a latent representation of the layer below. Fig. \ref{fig:CAE} plots the SCAE architecture for our $K$ = 2 case.

\noindent \textbf{Training Details} We first train the SCAE on both synthetic and real-world data in a unsupervised way, with a learning rate of 0.01 (we do not anneal it through training). Mean Squared Error (MSE) is used as the loss function. After SCAE is learned, its Conv. Layers 1 and 2 are imported to the CNN in Fig. \ref{fig:CNN}, as the $\mathbf{C_u}$ sub-network and fixed. The $\mathbf{C_s}$ sub-network, based on the output by $\mathbf{C_u}$, is then trained in a supervised manner. We start with the learning rate at 0.01, and follow a common heuristic to manually divide the learning rate by 10 when the validation error rate stops decreasing with the current rate. The ``dropout'' technique is applied to fc6 and fc7 layers during training.  Both $\mathbf{C_u}$ and $\mathbf{C_s}$ are trained with a default batch size of 128, momentum of 0.9 and weight decay of 0.0005. The network training is implemented using the CUDA ConvNet package \cite{imagenet}, and runs on a workstation with 12 Intel Xeon 2.67GHz CPUs and 1 GTX680 GPU. It takes around 1 day to complete the entire training pipeline. 

\noindent \textbf{Testing Details} We adopt multi-scale multi-view testing to improve the result robustness. For each test image, it is first normalized to 105 pixels in height, but squeezed in width by three different random ratios, all drawn from a uniform distribution between 1.5 and 3.5, matching the effects of squeezing and variable aspect ratio operations during training. Under each squeezed scale, five $105 \times 105$ patches are sampled at different random locations. That constitutes in total fifteen test patches, each of which comes with different aspect ratios and views, from one test image. As every single patch could produce a softmax vector through the trained CNN, we average all fifteen softmax vectors to determine the final classification result of the test image.

\subsection{Connections to Previous Work}
We are not the first to look into an essentially ``hierarchical'' deep architecture for domain adaption. In \cite{Ng}, the proposed transfer learning approach relies on the unsupervised learning of representations. Bengio et. al hypothesized in \cite{Bengio09} that more levels of representation can give rise to more abstract, more general features of the raw input, and that the lower layers of the predictor constitute a hierarchy of features that can be \textbf{shared} across variants of the input distribution. The authors in \cite{sentiment} used data from the union of all domains to learn their shared features, which is different from many previous domain adaptation methods that focus on learning features in a unsupervised way from the target domain only. However, their entire network hierarchy is learned in a unsupervised fashion, except for a simple linear classier trained on top of the network, i.e., $K=N-1$. In \cite{Ngtext}, the CNN learned a set of filters from raw images as the first layer, and those low-level filters are fixed when training higher layers of the same CNN, i.e., $K=1$. In other words, they either adopt a simple feature extractor ($K=1$), or apply a shallow classifier ($K=N-1$). Our CNN decomposition is different from prior work in that:
\begin{itemize}
\item  Our feature extractor $\mathbf{C_u}$ and classier $\mathbf{C_s}$ are both deep sub-networks with more than one layer (both $K$ and $N-K$ are larger than 1), which means that both are able to perform more sophisticated learning. More evaluations can be found in Section 5.2.
\item We learn ``shared-feature'' convolutional filters rather than fully-connected networks such as in \cite{sentiment}, the former of which is more suitable for visual feature extractions. 
\end{itemize}
The domain mismatch between synthetic and real-world data on the lower-level statistics can occur in more scenarios, such as real-world face recognition from rendered images or sketches, recognizing characters in real scenes with synthetic training, human pose estimation with synthetic images generated from 3D human body models. We conjecture that our framework can be applicable to those scenarios as well, where labeled real-world data is scarce but synthetic data can be easily rendered.

\section{Learning-based Model Compression}

The architecture in Fig. \ref{fig:CNN} contains a huge number of parameters. It is widely known that the deep models are heavily over-parameterized \cite{denil2013predicting} and thus those parameters can be compressed to reduce storage by exploring their structure. For a typical CNN, about 90\% of the storage is taken up by the dense connected layers, which shall be our focus for mode compression. 

\begin{figure}[htbp]
\centering
\begin{minipage}{0.235\textwidth}
\centering \subfigure[Standard scale] {
\includegraphics[width=\textwidth]{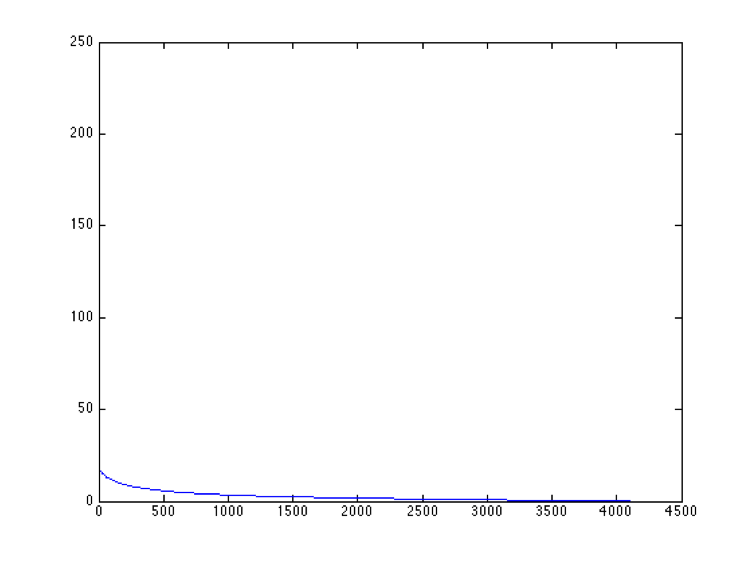}
}\end{minipage}
\begin{minipage}{0.235\textwidth}
\centering \subfigure[Logarithm scale] {
\includegraphics[width=\textwidth]{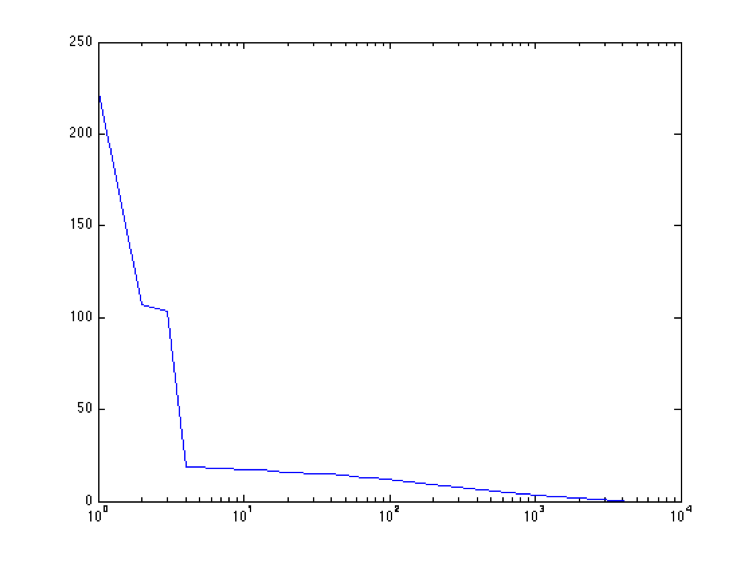}
}\end{minipage}
\caption{The plots of eigenvalues for the fc6 layer weight matrix in Fig. \ref{fig:CNN}. This densely connected layer takes up 85\% of the total model size.}
\label{rank}
\end{figure}

One way to shrink the number of parameters is using matrix factorization \cite{denton2014exploiting}. Given the parameter $W \in R^{m \times n}$,  we factorize it using singular-value decomposition (SVD):
\begin{equation}
\begin{array}{l}\label{SVD}
W=USV^T
\end{array}
\end{equation}
where $U \in R^{m \times m}$ and $V \in R^{n \times n}$ are two dense orthogonal matrices and $S \in R^{m \times n}$ is a diagonal matrix. To restore an approximate $W$ , we can utilize $\widetilde{U}$, $\widetilde{V}$ and $\widetilde{S}$, which denote the submatrices corresponding to the top $k$ singular vectors in $U$ and $V$ along with the top $k$ eigenvalue in $S$:
\begin{equation}
\begin{array}{l}\label{SVDapprox}
\widetilde{W}=\widetilde{U}\widetilde{S}\widetilde{V}^T
\end{array}
\end{equation}
The compression ratio given $m$, $n$, and $k$ is $\frac{k(m + n + 1)}{mn}$, which is very promising when $m, n \gg k$. However, the approximation of SVD is controlled by the decay along the eigenvalues in $S$. Even it is verified in Fig. \ref{rank} that eigenvalues of weight matrices usually decay fast (the 6-th largest eigenvalue is already less than 10\% of the largest one in magnitude), the truncation inevitably leads to information loss, and potential performance degradations, compared to the uncompressed model. 

Instead of first training a model then lossy-compressing its parameters, we propose to directly learn a \textbf{losslessly compressible} model (the term ``lossless'' is referred as there is no further loss after a model is trained). Assuming the parameter matrix $W$ of a certain network layer, our goal is to make sure that its rank is \textbf{exactly no more than a small constant $k$}. In terms of implementation, in each iteration, an extra hard thresholding operation \cite{ADM} is executed on $W$ after it is updated by a conventional back propagation step:
 \begin{equation}
\begin{array}{l}\label{hard}
W_k=U\mathcal{T}_k(S)V^T
\end{array}
\end{equation}
where $\mathcal{T}_k$ will keep the largest $k$ eigenvalues in $S$ while setting others to zeros. $W_k$ is best rank-$k$ approximation of $W$, as similarly in (\ref{SVDapprox}). However, different from (\ref{SVDapprox}), the proposed method incorporates low-rank approximation into model training and jointly optimize them as a whole, guaranteeing a rank-$k$ weight matrix that is ready to be compressed losslessly by applying (\ref{SVD}). Note there are other alternatives, such as vector quantization methods \cite{gong2014compressing}, that have been applied to compressing deep models with appealing performances. We will investigate utilizing them together to further compress our model in the future.

\section{Experiments}

\subsection{Analysis of Domain Mismatch}

We first analyze the domain mismatch between synthetic and real-world data, and examine how our synthetic data augmentation can help. First we define five dataset variations generated from VFR\_syn\_train and VFR\_real\_u. These are denoted by the letters N, S, F, R and FR and are explained in Table \ref{five}.

\begin{table*}[t]
\begin{center}
\caption{Comparison of Training and Testing Errors (\%) of Five SCAEs ($K$ = 2)}
\label{five}
\vspace{0.1em}
\begin{tabular}{|c|c|c|c|c|}
\hline
Methods & Training Data &Train & \multicolumn{2}{|c|}{Test}\\\cline{4-5}
 && & N & R \\
\hline
SCAE N  & {\bf N}: VFR\_syn\_train, no data augmentation & 0.02 & 3.54 & 31.28  \\
\hline
SCAE S  & {\bf S}: VFR\_syn\_train, standard augmentation 1-4 & 0.21 & 2.24 & 19.34  \\
\hline
SCAE F  & {\bf F}: VFR\_syn\_train, full augmentation 1-6 & 1.20 & 1.67 & 15.26\\
\hline
SCAE R  & {\bf R}:VFR\_real\_u, real unlabeled dataset & 9.64 &  5.73 & 10.87 \\
\hline
SCAE FR  & {\bf FR}: Combination of data from  {\bf F} and  {\bf R} &  6.52 &  2.02 &14.01\\
\hline
\end{tabular}
\end{center}
\end{table*}

We train five separate SCAEs, all of the same architecture as in Fig. \ref{fig:CAE}, using the above five training data variants. The training and testing errors are all measured by relative MSEs (normalized by the total energy) and compared in Table 1. The testing errors are evaluated on both the unaugmented synthetic dataset N and the real-world dataset R. Ideally, the better the SCAE captures the features from a domain, the smaller the reconstruction error will be on that domain.


As revealed by the training errors, real-world data contains rich visual variations and is more difficult to fit. The sharp performance drop from N to R of SCAE N indicates that the convolutional features for synthetic and real data are quite different. This gap is reduced in SCAE S, and further in SCAE F, which validates the effectiveness of adding font-specific data augmentation steps. SCAE R fits the real-world data best, at the expense of a larger error on N. SCAE FR achieves an overall best reconstruction performance of both synthetic and real-world images.

\begin{figure}[h]
\centering
\begin{minipage}{0.10\textwidth}
\centering \subfigure[original] {
\includegraphics[width=\textwidth]{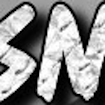}
}\end{minipage}
\begin{minipage}{0.10\textwidth}
\centering \subfigure[SCAE N] {
\includegraphics[width=\textwidth]{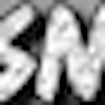}
}\end{minipage}
\begin{minipage}{0.10\textwidth}
\centering \subfigure[SCAE S] {
\includegraphics[width=\textwidth]{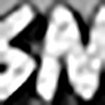}
}\end{minipage}
\\
\begin{minipage}{0.10\textwidth}
\centering \subfigure[SCAE F] {
\includegraphics[width=\textwidth]{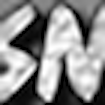}
}\end{minipage}
\begin{minipage}{0.10\textwidth}
\centering \subfigure[SCAE R] {
\includegraphics[width=\textwidth]{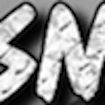}
}\end{minipage}
\begin{minipage}{0.10\textwidth}
\centering \subfigure[SCAE FR] {
\includegraphics[width=\textwidth]{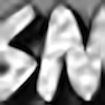}
}\end{minipage}
\caption{A real-world patch, and its reconstruction results from the five SCAE models.}
\label{areal} 
 \vspace{-0.5em}
\end{figure}

Fig. \ref{areal} shows an example patch from a real-world font image of highly textured characters, and its reconstruction outputs from all five models. The gradual visual variations across the results confirm the existence of a mismatch between synthetic and real-world data, and verify the benefit of data augmentation as well as learning shared features.



\subsection{Analysis of Network Structure}

\noindent \textbf{Fixing Network Depth $N$.} Given a fixed network complexity ($N$ layers), one may ask about how to best decompose the hierarchy to maximize the overall classification performance on real-world data. Intuitively, we should have sufficient layers of lower-level feature extractors as well as enough subsequent layers for good classification of labeled data. Thus, the depth $K$ of $C_u$ should neither be too small nor too large.

\begin{table}[h]
\begin{center}
\caption{Top-5 Testing Errors (\%) for Different CNN Decompositions (Varying $K$, $N$ = 8)}
\label{decom}
 \vspace{-1em}
\begin{tabular}{|c|c|c|c|c|c|c|}
\hline
K  &  0 & 1& 2 & 3 & 4 & 5\\
\hline
Train & 8.46 & 9.88 & 11.23 & 12.54 & 15.21 & 17.88  \\
\hline
VFR\_real\_test & 20.72 & 20.31 & 18.21 & 18.96 & 22.52 &25.97  \\
\hline
\end{tabular}
\end{center}
 \vspace{-0.6em}
\end{table}

\begin{figure}[htbp]
\centering
\begin{minipage}{0.10\textwidth}
\centering \subfigure[$K$=1] {
\includegraphics[width=\textwidth]{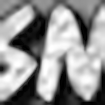}
}\end{minipage}
\begin{minipage}{0.10\textwidth}
\centering \subfigure[$K$=2] {
\includegraphics[width=\textwidth]{SR.png}
}\end{minipage}
\begin{minipage}{0.10\textwidth}
\centering \subfigure[$K$=4] {
\includegraphics[width=\textwidth]{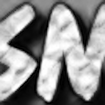}
}\end{minipage}
\begin{minipage}{0.10\textwidth}
\centering \subfigure[$K$=5] {
\includegraphics[width=\textwidth]{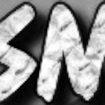}
}\end{minipage}
\caption{The reconstruction results of a real-world patch using SCAE FR, with different $K$ values.}
\label{fig:patch0}
\end{figure}

Table \ref{decom} shows that while the \textit{classification} training error increases with $K$, the testing error does not vary monotonically. The best performance is obtained with $K$ = 2 (3 slightly worse), where smaller or larger values of $K$ give substantially worse performance. When $K$ = 5, all layers are learned using SCAE, leading to the worst results. Rather than learning all hidden layers by unsupervised training, as suggested in \cite{sentiment} and other DL-based transfer learning work, our CNN decomposition reaches its optimal performance when higher-layer convolutional filters are still trained by supervised data. A visual inspection of reconstruction results of a real-world example in Fig. \ref{fig:patch0}, using SCAE FR with different $K$ values, shows that a larger $K$ causes less information loss during feature extraction and leads to a better reconstruction. But in the meantime, the classification result may turn worse since noise and irrelevant high frequency details (e.g. textures) might hamper recognition performance. The optimal $K$ =2 corresponds to a proper ``content-aware'' smoothening, filtering out ``noisy'' details while keeping recognizable structural properties of the font style.

\noindent \textbf{Fixing $C_s$ or $C_u$ Depth.} We investigate the influences of $K$ (the depth of $C_u$) when the depth of $C_s$ (e.g. $N-K$) keeps fixed. Table \ref{decomK} reveals that a deeper $C_u$ contributes little to the results. Similar trends are observed when we fix $K$ and adjust $N$ (and thus the depth of$C_s$). Therefore, we choose $N$= 8, $K$=2 to be the default setting.

\begin{table}[t]
\begin{center}
\caption{Top-5 Testing Errors (\%) for Different CNN Decompositions (Varying $K$, $N$ = $K$ + 6)}
\vspace{0.1em}
\label{decomK}
\begin{tabular}{|c|c|c|c|c|}
\hline
K  & 1& 2 & 3 & 4 \\
\hline
Train & 11.46 & 11.23 & 10.84 & 10.86   \\
\hline
VFR\_real\_test & 21.58 & 18.21 & 18.15 & 18.24   \\
\hline
\end{tabular}
\end{center}
 \vspace{-0.6em}
\end{table}

\subsection{Recognition Performances on VFR Datasets}

We implemented and evaluated the local feature embedding-based algorithm (LFE) in \cite{LFE} as a baseline, and include the four different DeepFont models as specified in Table \ref{full}. The first two models are trained in a fully supervised manner on $\mathbf{F}$, without any decomposition applied. For each of the later two models, its corresponding SCAE (SCAE FR for DeepFont CAE\_FR, and SCAE R for DeepFont CAE\_R) is first trained and then exports the first two convolutional layers to $\mathbf{C_u}$. All trained models are evaluated in term of top-1 and top-5 classification errors, on the VFR\_syn\_val dataset for validation purpose.  Benefiting from large learning capacity, it is clear that DeepFont models fit synthetic data significantly better than LFE. Notably, the top-5 errors of all DeepFont models (except for DeepFont CAE\_R) reach zero on the validation set, which is quite impressive for such a fine-grain classification task.

We then compare DeepFont models with LFE on the original VFRWild325 dataset in \cite{LFE}. As seen from Table \ref{full}, while DeepFont S fits synthetic training data best, its performance is the poorest on real-world data, showing a severe overfitting. With two font-specific data augmentations added in training, the DeepFont F model adapts better to real-world data, outperforming LFE by roughly 8\% in top-5 error. An additional gain of 2\% is obtained when unlabeled real-world data is utilized in DeepFont CAE\_FR. Next, the DeepFont models are evaluated on the new VFR\_real\_test dataset, which is more extensive in size and class coverage. A large margin of around 5\% in top-1 error is gained by DeepFont CAE\_FR model over the second best (DeepFont F), with its top-5 error as low as 18.21\%. We will use DeepFont CAE\_FR as the default DeepFont model.

Although SCAE R has the best reconstruction result on real-world data on which it is trained, it has large training and testing errors on synthetic data. Since our supervised training relies fully on synthetic data, an effective feature extraction for synthetic data is also indispensable. The error rates of DeepFont CAE\_R are also worse than those of DeepFont CAE\_FR and even DeepFont F on the real-world data, due to the large mismatch between the low-level and high-level layers in the CNN.

\begin{figure}[htbp]
\centering
\begin{minipage}{0.40\textwidth}
\centering {
\includegraphics[width=\textwidth]{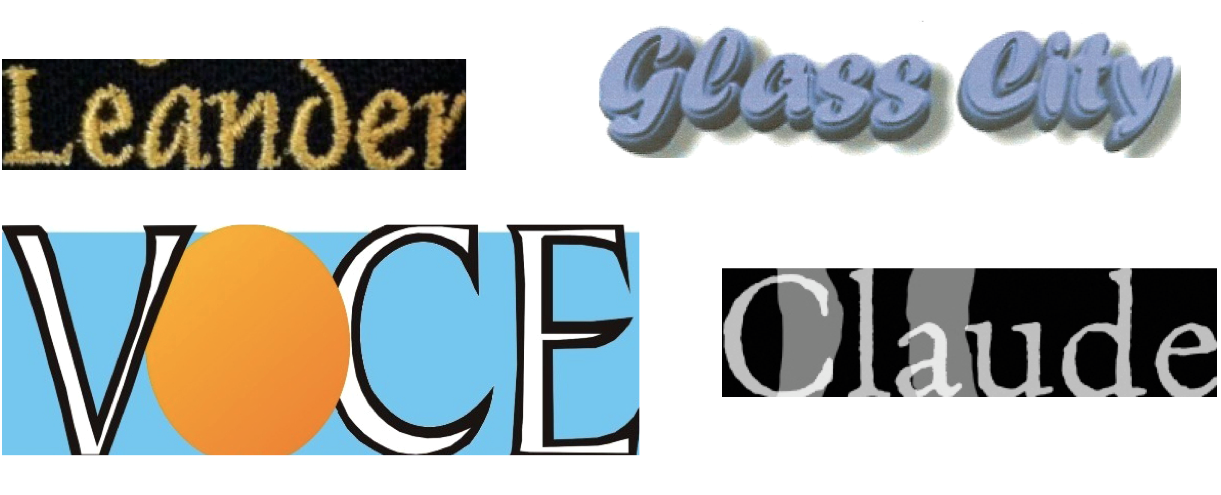}
}\end{minipage}
\caption{Failure VFR examples using DeepFont.}
\label{fig:fail}
\end{figure}

Another interesting observation is that all methods get similar top-5 errors on VFRWild325 and VFR\_real\_test, showing their statistical similarity. However, the top-1 errors of DeepFont models on VFRWild325 are significantly higher than those on VFR\_real\_test, with a difference of up to 10\%. In contrast, the top-1 error of LFE rises more than 13\% on VFR\_real\_test than on VFRWild325. For the small VFRWild325, the recognition result is easily affected by ``bad" examples (e.g, low resolution or highly compressed images) and class bias (less than 4\% of all classes are covered). On the other hand, the larger VFR\_real\_test dataset dilutes the possible effect of outliers, and examines a lot more classes. 


\begin{table*}[htb]
\vspace{-0.6em}
\begin{center}
\caption{Comparison of Training and Testing Errors on Synthetic and Real-world Datasets (\%)}
\label{full}
\vspace{0em}
\begin{tabular}{|c|c|c|c|c|c|c|c|c|c|}
\hline
Methods & \multicolumn{2}{|c|}{Training Data} & Training & \multicolumn{2}{|c|}{VFR\_syn\_val} & \multicolumn{2}{|c|}{VFRWild325} &\multicolumn{2}{|c|}{VFR\_real\_test}\\\cline{2-3}
\cline{5-10}
&  $\mathbf{C_u}$ & $\mathbf{C_s}$ & Error &  Top-1 & Top-5 &  Top-1 & Top-5 &  Top-1 & Top-5 \\
\hline
LFE   & /\  & /\  & /\  & 26.50 & 6.55 & 44.13 & 30.25 & 57.44 &  32.69 \\
\hline
DeepFont S   & /\ & F & 0.84 & 1.03 & 0 & 64.60 & 57.23 & 57.51 & 50.76 \\
\hline
DeepFont F   & /\ & F & 8.46 & 7.40 & 0  & 43.10 & 22.47 & 33.30 & 20.72 \\
\hline
DeepFont CAE\_FR  & FR & F & 11.23 & 6.58 & 0 & {\bf 38.15} & {\bf 20.62} & {\bf 28.58} & {\bf 18.21}  \\
\hline
DeepFont CAE\_R & R  & F & 13.67 & 8.21 & 1.26 & 44.62 & 29.23 & 39.46 & 27.33  \\
\hline
\end{tabular}
\end{center}
\end{table*}

\begin{figure}[htbp]
\centering
\begin{minipage}{0.42\textwidth}
\centering \subfigure[] {
\includegraphics[width=\textwidth]{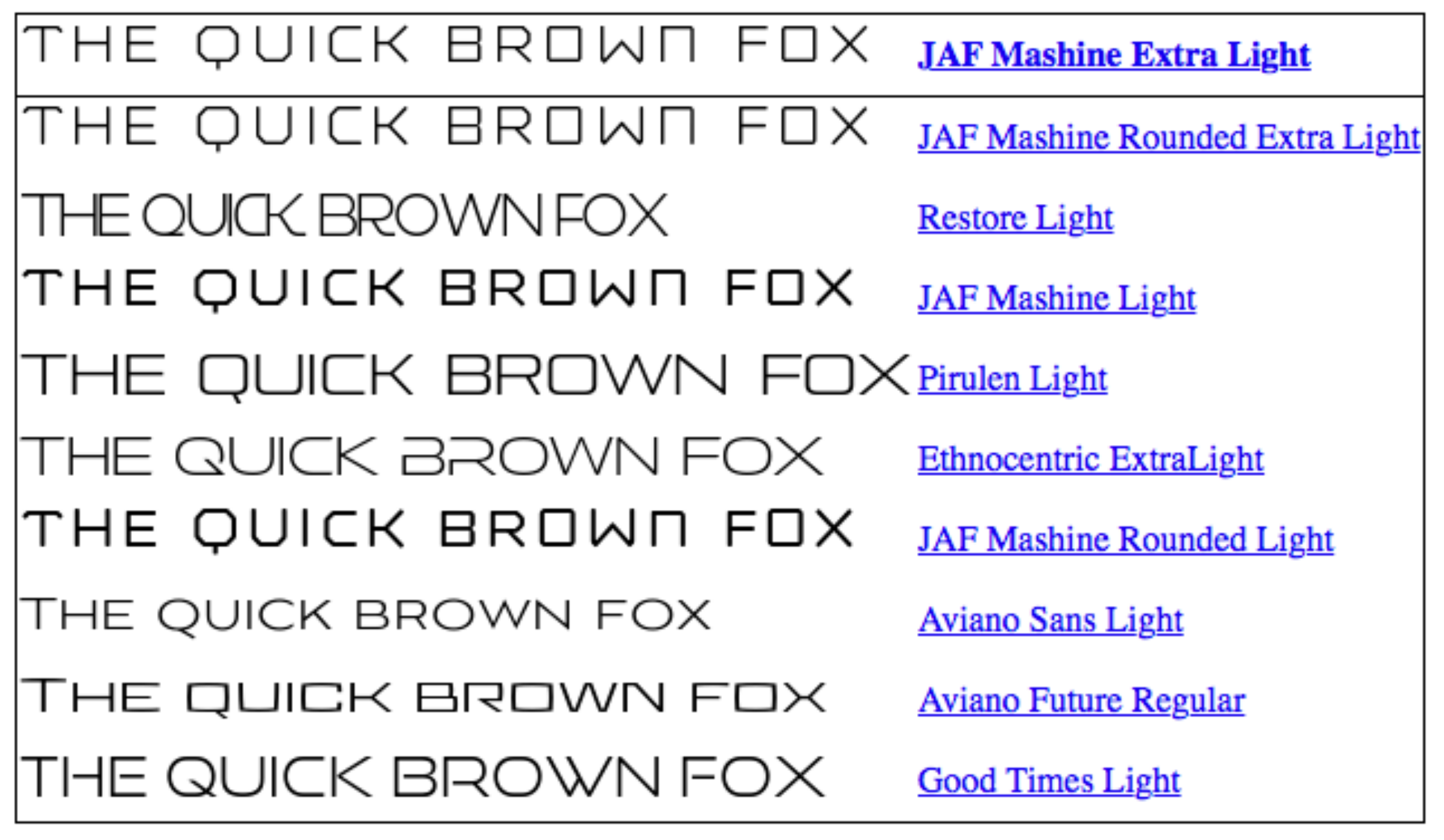}
}\end{minipage}
\begin{minipage}{0.42\textwidth}
\centering \subfigure[] {
\includegraphics[width=\textwidth]{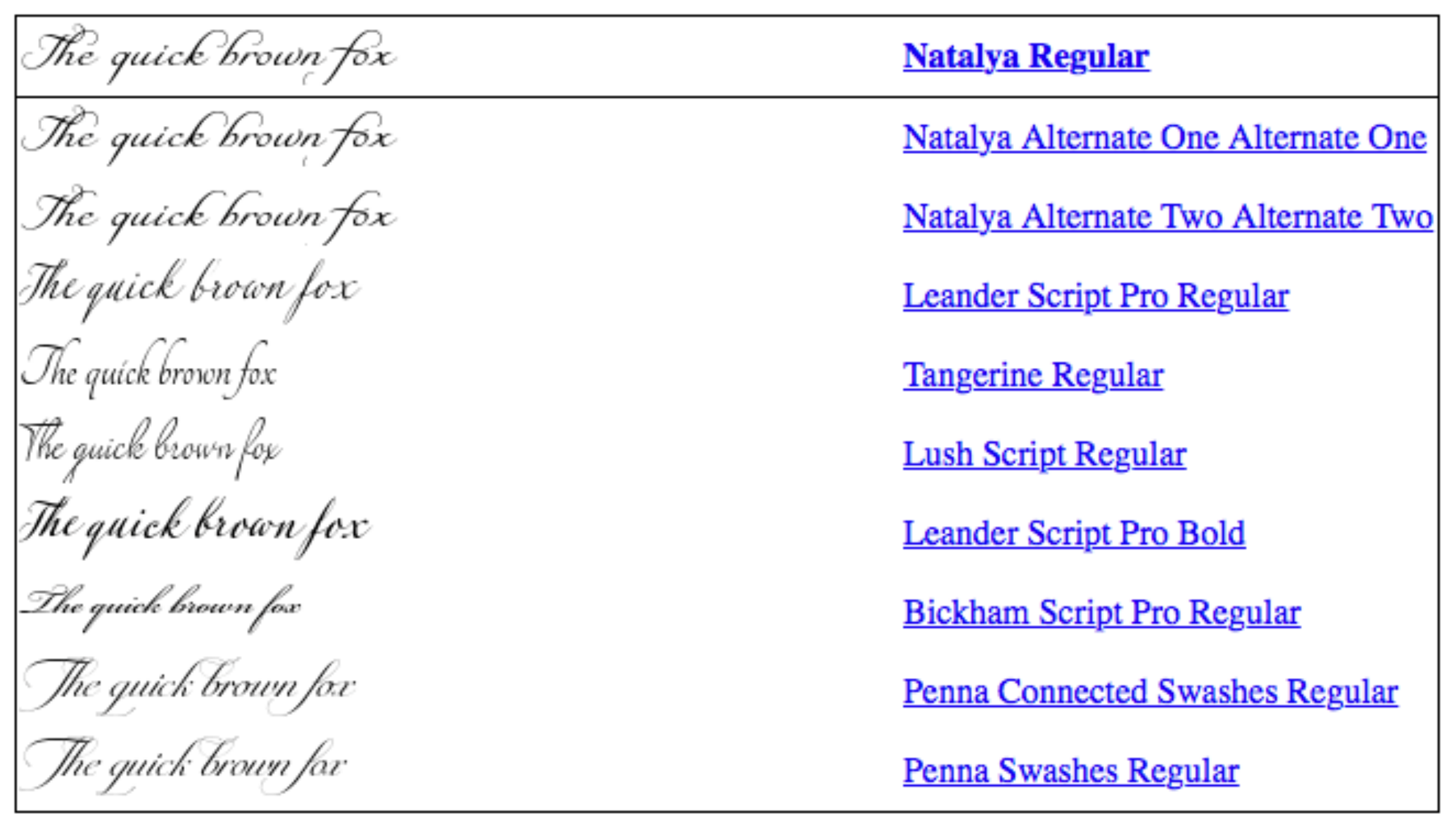}
}\end{minipage}
\begin{minipage}{0.42\textwidth}
\centering \subfigure[] {
\includegraphics[width=\textwidth]{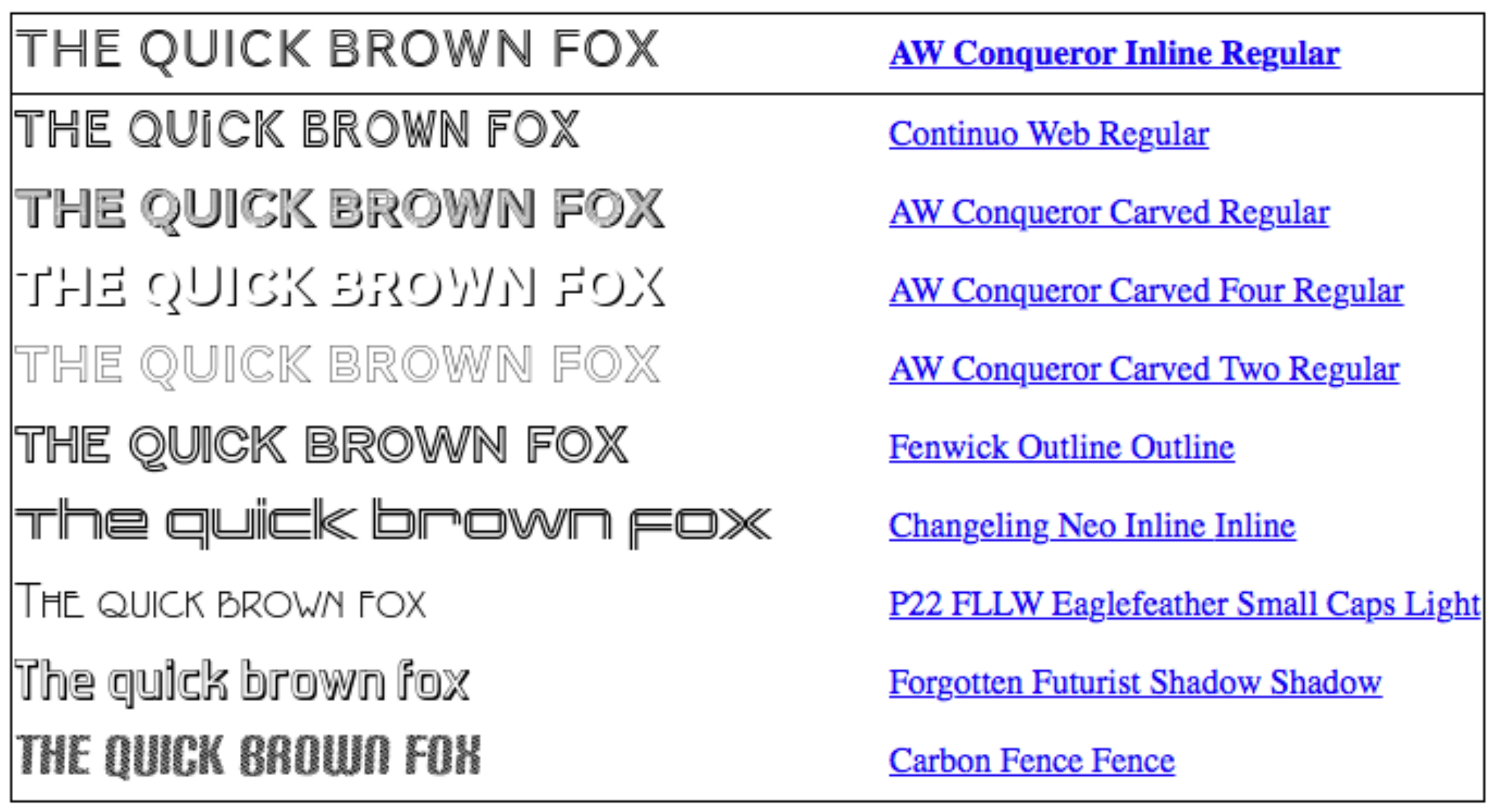}
}\end{minipage}
\caption{Examples of the font similarity. For each one, the top is the query image, and the renderings with the most similar fonts are returned.}
\label{fig:graph}
\end{figure}

Fig. \ref{fig:fail} lists some failure cases of DeepFont. For example, the top left image contains extra ``fluff'' decorations along text boundaries, which is nonexistent in the original fonts, that makes the algorithm incorrectly map it to some ``artistic'' fonts. Others are affected by 3-D effects, strong obstacles in foreground, and in background. Being considerably difficult to be adapted, those examples fail mostly because there are neither specific augmentation steps handling their effects, nor enough examples in VFR\_real\_u to extract corresponding robust features.

\subsection{Evaluating Font Similarity using DeepFont}

There are a variety of font selection tasks with different goals and requirements. One designer may wish to match a font to the style of a particular image. Another may wish to find a free font which looks similar to a commercial font such as Helvetica. A third may simply be exploring a large set of fonts such as Adobe TypeKit or Google Web Fonts. Exhaustively exploring the entire space of fonts using an alphabetical listing is unrealistic for most users. The authors in \cite{exploratory} proposed to select fonts based on online crowdsourced attributes, and explore \textit{font similarity}, from which a user is enabled to explore other visually similar fonts given a specific font. The font similarity measure is  very helpful for font selection, organization, browsing, and suggestion. 

Based on our DeepFont system, we are able to build up measures of font similarity.  We use the $4096 \times 1$ outputs of the fc7 layer as the high-level feature vectors describing font visual appearances. We then extract such features from all samples in VFR\_syn\_val Dataset, obtaining 100 feature vectors per class. Next for each class, the 100 feature vectors is averaged to a representative vector. Finally, we calculate the Euclidean distance between the representative vectors of two font classes as their similarity measure. Visualized examples are demonstrated in Fig. \ref{fig:graph}. For each example, the top is the query image of a known font class; the most similar fonts obtained by the font similarity measures are sorted below. Note that although the result fonts can belong to different font families from the query, they share identifiable visual similarities by human perception. 

Although not numerically verified as in \cite{exploratory}, the DeepFont results are qualitatively better when we look at the top-10 most similar fonts for a wide range of query fonts. The authors of \cite{exploratory} agree per personal communication with us.

\subsection{DeepFont Model Compression}

Since the fc6 layer takes up 85\% of the total model size, we first focus on its compression. We start from a well-trained DeepFont model (DeepFont CAE\_FR), and continue tuning it with the hard thresholding (\ref{hard}) applied to the fc6 parameter matrix $W$ in each iteration, until the training/validation errors reach the plateau again. 

Table \ref{compress} compares the DeepFont models compressed using conventional matrix factorization (denoted as the ``lossy'' method), and the proposed learning based method (denoted as the ``lossless'' method), under different compression ratios (fc6 and total size counted by parameter numbers). The last column of Table \ref{compress} lists the top-5 testing errors (\%) on VFR\_real\_test. We observe a consistent margin of the ``lossless'' method over its ``lossy'' counterpart, which becomes more significant when the compression ratio goes low (more than 1\% when $k$ = 5). Notably, when $k$ = 100, the proposed ``lossless'' compression suffers no visible performance loss, while still maintaining a good compression ratio of 5.79.

\begin{table}[ht]
\begin{center}
\caption{Performance Comparisons of Lossy and Lossless Compression Approaches}
\vspace{-1em}
\label{compress}
\begin{tabular}{|c|c|c|c|c|c|}
\hline
  & fc6 size & Total size & Ratio & \textbf{Method} &  Error  \\
\hline
default & \scriptsize{150,994,944} & \scriptsize{177,546,176} & NA & NA & 18.21   \\
\hline
\multirow{2}{*}{$k$=5} & \multirow{2}{*}{\scriptsize{204,805}} & \multirow{2}{*}{\scriptsize{26,756,037}} & \multirow{2}{*}{6.64} &  Lossy & 20.67  \\
\cline{5-6}
& &  & & Lossless & 19.23   \\
\hline
\multirow{2}{*}{$k$=10} & \multirow{2}{*}{\scriptsize{409,610}} & \multirow{2}{*}{\scriptsize{26,960,842}} & \multirow{2}{*}{6.59} & Lossy & 19.25   \\
\cline{5-6}
& &  & & Lossless & 18.87   \\
\hline
\multirow{2}{*}{$k$=50} & \multirow{2}{*}{\scriptsize{2,048,050}} & \multirow{2}{*}{\scriptsize{28,599,282}} & \multirow{2}{*}{6.21} & Lossy  & 19.04   \\
\cline{5-6}
& &  & & Lossless & 18.67   \\
\hline
\multirow{2}{*}{$k$=100} & \multirow{2}{*}{\scriptsize{4,096,100}}  & \multirow{2}{*}{\scriptsize{30,647,332}} & \multirow{2}{*}{5.79} & Lossy & 18.68   \\
\cline{5-6}
& &  & & Lossless & 18.21   \\
\hline
\end{tabular}
\end{center}
 \vspace{-0.6em}
\end{table}

In practice, it takes around 700 megabytes to store all the parameters in our uncompressed DeepFont model, which is quite huge to be embedded or downloaded into most customer softwares. More aggressively, we reduce the output sizes of both fc6 and fc7 to 2048, and further apply the proposed compression method ($k$ = 10) to the fc6 parameter matrix. The obtained ``mini'' model, with only 9, 477, 066 parameters and a high compression ratio of 18.73, becomes less than 40 megabytes in storage. Being portable even on mobiles, It manages to keep a top-5 error rate around 22\%.

\section{Conclusion}

In the paper, we develop the DeepFont system to remarkably advance the state-of-the-art in the VFR task. A large set of labeled real-world data as well as a large corpus of unlabeled real-world images is collected for both training and testing, which is the first of its kind and will be made publicly available soon. While relying on the learning capacity of CNN, we need to combat the mismatch between available training and testing data. The introduction of SCAE-based domain adaption helps our trained model achieve a higher than 80\% top-5 accuracy. A novel lossless model compression is further applied to promote the model storage efficiency. The DeepFont system not only is effective for font recognition, but can also produce a font similarity measure for font selection and suggestion.

\bibliographystyle{abbrv}
\normalsize
\bibliography{sigproc}

\end{document}